\documentclass[10pt]{article}
\usepackage[letterpaper]{geometry}
\usepackage{hicss}
\usepackage{times}
\usepackage[none]{hyphenat}
\usepackage{url}
\usepackage{latexsym}
\usepackage{minted}
\usepackage{indentfirst}
\usepackage{graphicx}
\graphicspath{{images/}}
\usepackage[
    style=apa,
  ]{biblatex}
\usepackage{booktabs}
\usepackage{hyperref}
\usepackage{amsmath}

\usepackage{caption}
\title{Detecting GPT-Assisted Writing Using Interpretable Stylometric Features}

\author{
Rajesh Kumar \\
Bucknell University \\
{\underline{rajesh.kumar@bucknell.edu}}
\And
Nabeel Siddiqui \\
Susquehanna University \\
{\underline{siddiqui@susqu.edu}}
\And
Alexander Fuchsberger \\
Bucknell University \\
{\underline{af033@bucknell.edu}}
}

\date{today}

\begin{document}
\maketitle
\begin{abstract}
Distinguishing GPT-assisted from independently authored student writing has become a critical challenge in academia. This paper evaluates the discriminative capability of interpretable stylometric features extracted solely from submitted text. Using data from $90$ participants who wrote both independently and with ChatGPT assistance, we evaluate eight machine learning classifiers while keeping data from the same participant together during validation. On the held-out test set, Random Forest achieved an ROC-AUC of $0.870$ and an F1-score of $0.842$, with False Positive and False Negative rates of $22.2\%$ and $11.1\%$, respectively. SHAP analysis shows that lexical and grammatical characteristics drive the resulting predictions. The findings suggest that transparent, text-intrinsic features provide measurable signal for detecting GPT-assisted writing.
\end{abstract}

\subsubsection*{Keywords: AI text detection, stylometry, interpretable machine learning}

\section{Introduction}
\label{sec:introduction}
Large Language Models (LLMs) have changed how students produce written assignments. While they can provide personalized tutoring, on-demand explanations, brainstorming assistance, and writing support, they also raise concerns regarding authorship and academic integrity \parencite{unesco2023chatgpt,Kundu2024,Roh2025,mehta2025}. Recent evidence indicates that student AI use in education has become routine rather than exceptional. In the Lumina Foundation–Gallup 2026 State of Higher Education Study, $57\%$ of currently enrolled U.S. college students reported using AI in their coursework at least weekly, including about one in five who reported using it daily. More than half reported using AI daily or weekly to edit or improve their writing, and $36\%$ reported using it daily or weekly to write papers \parencite{gallup2026}. As GPT-assisted writing becomes increasingly integrated into educational workflows, institutions require reliable methods for distinguishing independently authored and GPT-assisted student work.
 
This need has led to rapid growth in AI text detection research. Existing approaches include statistical detection and visualization tools such as GLTR \parencite{gehrmann2019gltr}, zero-shot detection methods such as DetectGPT \parencite{mitchell2023detectgpt}, watermarking approaches that embed detectable signals during text generation \parencite{kirchenbauer2023watermark}, adversarially trained detectors such as RADAR \parencite{hu2023radar}, and fine-grained detection systems including LLM-DetectAIve \parencite{abassy2024llm}. Recent studies have also explored behavioral approaches based on keystroke dynamics \parencite{Kundu2024,Roh2025,mehta2025}. Although these techniques have reported reasonable performance, recent studies have questioned detector robustness under recursive paraphrasing \parencite{sadasivan2023can} and stylistic or text-complexity variation \parencite{doughman2024exploring}.

A key challenge is detecting GPT-assisted writing while also providing an additional layer of evidence that instructors can examine during academic integrity investigations. Misconduct decisions may affect grades and disciplinary actions. Such decisions require explanations that instructors and review committees can examine, not opaque prediction scores from proprietary models whose internal evidence cannot be inspected. The explainable artificial intelligence literature frames explanations as a way to make model behavior understandable to humans, examine predictive models, and assess properties such as reliability, fairness, usability, and trust \parencite{doshivelez2017towards,biecek2021explanatory,ribeiro2016why}. Detection therefore needs approaches that characterize GPT-assisted writing through observable linguistic evidence rather than latent neural representations.

\emph{Stylometry} provides a natural foundation for such an approach. Rather than modeling semantic content, stylometric analysis measures writing through characteristics such as lexical diversity, vocabulary richness, grammatical composition, and sentence structure. Stylometry has a long history in authorship attribution, including the work of \textcite{mosteller1964inference} and subsequent extensions through stylistic distance measures \parencite{burrows2002delta}, authorship verification methods \parencite{koppel2004authorship}, and function-word classification approaches \parencite{koppel2002automatically}. Computational linguistics and stylometry have also studied measures of lexical richness and vocabulary diversity \parencite{tweedie1998how,mccarthy2010mtld}. 

Despite extensive research on AI text detection, prior work has given relatively little attention to whether a compact set of interpretable stylometric features extracted from the final submitted text alone could provide sufficient evidence for detection. This study investigates that question, focusing on text-intrinsic evidence that can be applied retrospectively without access to the writing process. The following are the main contributions:

\begin{itemize}

\item We investigate whether a set of interpretable stylometric features extracted from submitted text can distinguish GPT-assisted writing from independently authored student writing.

\item We propose a sliding window-based protocol (250-word windows with a 125-word stride) to reduce the impact of document length on the extracted features, along with disjoint sets of users across the training and testing partitions to prevent data leakage and evaluate performance on unseen users.

\item We systematically evaluate eight machine learning classifiers using ROC-AUC, F1, False Positive, and False Negative rates. Window-level probabilities are aggregated into document-level decisions using a pre-specified median-probability rule.

\item We examine the interpretability of Random Forest, which ranked first
during training-data validation, using SHAP analysis to identify the
stylometric features that contribute most strongly to its predictions.

\end{itemize}

The remainder of this paper is organized as follows. Section~\ref{sec:related-work} reviews prior research on AI-generated text detection and stylometric analysis. Section~\ref{sec:methods} describes the dataset, stylometric feature extraction process, machine learning models, and experimental methodology. Section \ref{sec:results} presents the empirical results, statistical analyses, and SHAP-based interpretation of Random Forest. Section~\ref{sec:ethical} discusses ethical considerations and limitations. Finally, Section~\ref{sec:conclusion} summarizes the main findings and outlines directions for future research.

\begin{figure*}[htp]
\centering
\includegraphics[width=4in, height=1.9in]{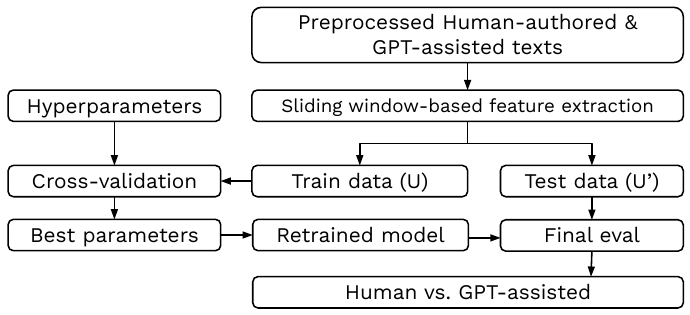}
\caption{Overview of the proposed detection framework. Stylometric features are extracted from fixed-length sliding windows, with disjoint sets of users used for training and testing. Hyperparameters are selected through cross-validation on the training set, and window-level predictions are aggregated using the median to obtain the final document-level classification.}
\label{fig:framework}
\end{figure*}

\section{Related Work}
\label{sec:related-work}

Existing detection approaches primarily infer authorship from statistical regularities learned by language models or introduced during text generation. DetectGPT, for example, exploits the observation that machine-generated text tends to occupy regions of negative probability curvature, enabling zero-shot detection without additional training \parencite{mitchell2023detectgpt}. Watermarking methods alter the generation process by embedding detectable signatures into token selection \parencite{kirchenbauer2023watermark}. Although these approaches offer reasonable evidence, they rely on proprietary models, access to generation probabilities, or high-dimensional neural representations that often give limited insight into the evidence supporting individual predictions. Recent studies have further reported reduced robustness under recursive paraphrasing \parencite{sadasivan2023can} and stylistic or text-complexity variation \parencite{doughman2024exploring}, as well as systematic biases against non-native English writers \parencite{liang2023gpt}, which raise concerns about detector use in high-stakes educational settings. 

Parallel research has shifted focus from the submitted document to the writing process. Keystroke dynamics captures observable behavioral signals, including typing speed, pauses, revisions, and temporal writing patterns, that can differ between independently authored and GPT-assisted writing \parencite{Kundu2024,Roh2025,mehta2025}. However, behavioral approaches require instrumented writing environments and therefore do not apply retrospectively to submitted assignments or existing document collections.

Stylometry offers a complementary text-intrinsic perspective based on the hypothesis that writers exhibit lexical and grammatical preferences that remain sufficiently consistent to characterize writing style. Authorship studies show that surface-level lexical and syntactic features carry enough signal to distinguish writers \parencite{mosteller1964inference,burrows2002delta,koppel2002automatically,koppel2004authorship}. Lexical diversity measures and vocabulary richness metrics capture aspects of writing style while remaining directly observable and statistically interpretable \parencite{tweedie1998how,mccarthy2010mtld}. We also include entropy as an information-theoretic summary of token-frequency dispersion \parencite{shannon1948mathematical}. Unlike proprietary prediction scores or latent neural embeddings, these features correspond to explicit linguistic characteristics that can be independently examined by educators and academic integrity investigators. While GPT can imitate human writing style, detectable differences may
remain \parencite{Rebira2025}. Another closely related study \textcite{safi2025detecting} used a transparent classification scheme based on word choices and function words to distinguish student responses from ChatGPT responses, and found that lightly edited or style-mimicking AI text remained detectable. We extend this line of work with evaluation on unseen participants, SHAP-based feature-level explanations, and explicit confidence intervals on error rates.

\section{Methodology}
\label{sec:methods}
Figure~\ref{fig:framework} describes the detection framework we implemented and evaluated. We first extract the features listed in Table~\ref{tab:features} using a sliding window-based protocol. We then create two disjoint sets of users, one for training the detectors and the other for evaluating their performance. Each detector produces a probability score for each window (sample), and the window-level probabilities are aggregated using a median-probability rule to obtain a document-level score. The document-level scores are then used to classify each document as human-authored or GPT-assisted. The performance of each detector on the test data is evaluated using F1, ROC-AUC, False Positive Rate, and False Negative Rate. Finally, we use SHAP to examine the stylometric features contributing to the predictions of Random Forest.
 
\begin{table*}[htp]
\centering
\caption{Stylometric representation used for GPT-assisted writing detection. Each writing sample is represented by nine interpretable lexical, syntactic, and sentence-level features that quantify vocabulary diversity, vocabulary richness, grammatical composition, and sentence-length variation.}
\label{tab:features}
\renewcommand{\arraystretch}{1.15}
\small
\begin{tabular}{@{}p{1.0cm} p{3.5cm} p{2.8cm} p{7.2cm}@{}}
\hline
\textbf{Feature} & \textbf{Name} & \textbf{Computation} & \textbf{Description} \\
\hline

TTR & Type-token ratio &
$\displaystyle |V|/N$ &
Proportion of unique tokens. \\

Hapax Ratio & Hapax legomena ratio &
$\displaystyle H/N$ &
Proportion of tokens occurring exactly once. \\

Word Entropy & Word entropy &
$\displaystyle -\sum_{i=1}^{|V|} p_i \log_2(p_i)$ &
Shannon entropy of the token frequency distribution. \\

NSR & Non-stopword ratio &
$\displaystyle S/N$ &
Proportion of non-stopword tokens. \\

NOUN Ratio & Noun ratio &
$\displaystyle \#NOUN/ N_{POS}$ &
Proportion of tokens tagged as nouns. \\

VERB Ratio & Verb ratio &
$\displaystyle \#VERB/N_{POS}$ &
Proportion of tokens tagged as verbs. \\

ADJ Ratio & Adjective ratio &
$\displaystyle \#ADJ/N_{POS}$ &
Proportion of tokens tagged as adjectives. \\

ADV Ratio & Adverb ratio &
$\displaystyle \#ADV/N_{POS}$ &
Proportion of tokens tagged as adverbs. \\

SLV & Sentence-length variability &
$\displaystyle \sqrt{\frac{\sum_{i=1}^{m}(L_i-\bar{L})^2}{m-1}}$ &
Standard deviation of sentence lengths. \\

\hline
\end{tabular}

\vspace{2pt}
\footnotesize{
$N$ = total number of tokens;
$|V|$ = number of unique tokens;
$H$ = number of hapax legomena;
$p_i$ = relative frequency of token $i$;
$S$ = number of non-stopword tokens;
$N_{POS}$ = total POS-tagged tokens;
$m$ = number of sentences;
$L_i$ = length of sentence $i$ in tokens;
$\bar{L}$ = mean sentence length.
}
\end{table*}

\subsection{Dataset}
We use the IIITD-BU (Paraphrased) dataset described by \textcite{mehta2025}. The dataset was collected to examine student writing under independent and GPT-assisted conditions using keystroke dynamics \parencite{mehta2025}. Participants were primarily students enrolled in a course on Large Language Models at IIIT Delhi, India. They responded to two questions about large language models: one focused on explaining how LLMs work and their applications, while the other asked participants to analyze their strengths and weaknesses in educational settings and propose an improvement.

Participants completed the tasks in two sessions. In the first session, they answered the questions independently using their own knowledge and reasoning. In the second session, they were allowed to use LLMs such as ChatGPT for assistance but were explicitly instructed to paraphrase the generated content before typing their responses. Copy-and-paste functionality was disabled, as were auto-correct, grammar tools, and browser extensions. Thus, in this study, ``GPT-assisted'' refers specifically to responses written by paraphrasing GPT-generated content rather than to the broader range of possible GPT-assisted writing practices.

Although the original study collected keystroke dynamics and interaction traces, we focus exclusively on the final submitted text. We make this choice deliberately because our objective is to determine whether interpretable stylometric features extracted directly from submitted text contain sufficient signal to distinguish the two writing conditions without
behavioral information, which may be unavailable in retrospective academic integrity investigations.

Each participant completed two prompts in each session. We concatenate the two responses from the same session to form a single response for each participant under each writing condition. The resulting dataset contains $180$ writing samples from 90 participants, with each participant contributing one independently authored response and one GPT-assisted response. The
dataset is therefore balanced across the two writing conditions.

Independently authored responses contained an average of $620$ whitespace-delimited words (SD = $198$), compared with $515$ words (SD = $155$) for GPT-assisted responses. The corresponding mean sentence counts were $26.5$ and $24.7$, while mean sentence lengths were $27.2$ and $27.3$ words, respectively. Because differences in document length could influence some stylometric features, particularly lexical-diversity measures, Section~\ref{sec:results} explicitly examines these differences and whether document length alone can explain the observed classification performance.

Because every participant appears in both conditions, a naive random split could place a participant's independently authored and ChatGPT-assisted texts on opposite sides of the split, allowing author-specific writing style to leak between training and evaluation and potentially inflating performance. To prevent this, we use disjoint sets of users for training and evaluation, with all texts from a given participant kept in the same set. The $90$ participants were partitioned into $72$ training participants ($144$ samples) and $18$ held-out participants ($36$ samples: $18$ independently authored and $18$ ChatGPT-assisted), with no participant shared between the two sets.

\subsection{Preprocessing and feature engineering}
We extracted the final text from each response and removed residual keyboard-event tokens from the original keystroke-logging instrumentation, including \texttt{Control} and \texttt{CapsLock}. We split each cleaned document into whitespace-delimited words and generated windows using a pre-specified length of $250$ words and a stride of $125$ words, resulting in
$50\%$ overlap between adjacent windows. We right-aligned the final window to the end of the document so that no text was discarded, and no artificial padding was added. Documents shorter than 250 words contributed one full-document window.

This procedure produced $722$ windows, including $397$ independently authored windows and $325$ GPT-assisted windows. The $72$ training participants contributed 595 windows, while the $18$ held-out participants contributed $127$ windows from $36$ documents, including $74$ independently authored and $53$ GPT-assisted windows. Windowing increases the number of text segments available to the classifiers, but these windows are not additional independent participants. Therefore, we kept all windows from a given participant within the same training, validation, or test set throughout the analysis.

We tokenized each window using \texttt{quanteda} with punctuation removed and obtained part-of-speech annotations using the \texttt{udpipe} English EWT model (\texttt{english-ewt-ud-2.5}). The Non-stopword Ratio used the 175-word Snowball English stop list. Stop-list matching was case-folded, while Type-Token Ratio, Hapax Ratio, and Word Entropy preserved case. We segmented sentences at sentence-final periods for Sentence-Length Variability. We computed the nine features independently for each window and centered and scaled them using parameters estimated from the training data only. All analyses used R 4.4.3 with \texttt{fastml} (0.7.8), \texttt{quanteda} (4.3.1), and \texttt{udpipe} (0.8.16). A fixed random seed (123) was used for the user split, cross-validation folds, and Bayesian tuning.

Each window received a predicted probability of GPT assistance. We aggregated these probabilities using the median to obtain a document-level score. Using a pre-specified threshold of $0.50$, we classified a document as GPT-assisted when its median probability was at least $0.50$. The median reduces the influence of an unusually high or low window while retaining the probability information that would be lost with hard majority voting.

\subsection{Feature analysis}
We analyzed the nine stylometric features to examine how they differ between independently authored and GPT-assisted writing. Because overlapping windows from the same document are not statistically independent, for this analysis we first summarized each feature by its median across the windows of each document. We then used paired $t$-tests to compare each feature between the two writing conditions across the $90$ participants. We also computed point-biserial correlations ($r$) to measure the association between each feature and the writing condition.

\begin{table}[htp]
\centering
\caption{Mean document-level median feature values for independently authored and GPT-assisted writing across 90 participants. Paired $t$-tests compare the two writing conditions, and positive $r$ indicates higher values in GPT-assisted writing.}
\label{tab:feature-tests}
\setlength{\tabcolsep}{4pt}
\begin{tabular}{lccrrr}
\toprule
\textbf{Feature} & \textbf{Human} & \textbf{AI} & \textbf{$t$} & \textbf{$p$} & \textbf{$r$} \\
\midrule
TTR         & 0.60  & 0.65  &  6.66 & .001 &  0.43 \\
Hapax Ratio  & 0.44  & 0.51  &  7.19 & .001 &  0.46 \\
Word Entropy & 6.74  & 6.94  &  7.98 & .001 &  0.42 \\
SLV           & 14.67 & 13.07 & -2.22 & .03  & -0.13 \\
NSR          & 0.59  & 0.64  &  8.80 & .001 &  0.50 \\
NOUN Ratio   & 0.24  & 0.27  &  8.37 & .001 &  0.47 \\
VERB Ratio     & 0.13  & 0.14  &  2.52 & .01  &  0.18 \\
ADJ Ratio      & 0.08  & 0.08  &  2.09 & .04  &  0.15 \\
ADV Ratio    & 0.05  & 0.04  & -6.21 & .001 & -0.34 \\
\bottomrule
\end{tabular}

\end{table}

\subsection{Classifier training and validation}
 
We evaluated Logistic Regression, Linear Discriminant Analysis, Quadratic Discriminant Analysis, Naive Bayes, Random Forest, linear Support Vector Machine, $k$-Nearest Neighbors, and a multilayer perceptron. These classifiers were selected because they represent different learning paradigms, including linear and discriminant models, probabilistic methods, instance-based learning, ensemble learning, margin-based classification, and neural networks. They are also commonly used in text classification and related stylometric classification tasks. Evaluating this diverse set of classifiers allows us to examine whether the proposed stylometric features are useful across different modeling approaches rather than being specific to a single classifier.

Hyperparameter tuning was performed using only the training partition through Bayesian optimization with 30 iterations over the default \texttt{fastml} search space for each classifier. During validation, all windows and both writing conditions from a given participant were kept in the same fold. We compared the eight classifiers using mean validation ROC-AUC under repeated $10$-fold validation and an additional 5-by-5 nested-validation check. Random Forest ranked first under both validation procedures. The final Random Forest used three candidate features at each split, $500$ trees, and a minimum node size of $10$.

After tuning, each classifier was fit to the complete training partition and evaluated once on the held-out test set. The held-out test set was not used for hyperparameter tuning or model selection.

\subsection{Evaluation metrics}

We treat GPT-assisted writing as the positive class and independently authored writing as the negative class. We use ROC-AUC and F1-score as the principal performance measures. ROC-AUC measures how well the document-level probabilities rank GPT-assisted texts above independently authored texts across different thresholds. F1 summarizes the balance between precision
and recall for detecting GPT-assisted writing.

We also report the False Positive Rate (FPR) and False Negative Rate (FNR). A false positive occurs when independently authored writing is incorrectly classified as GPT-assisted, while a false negative occurs when GPT-assisted writing is incorrectly classified as independently authored. We report both errors as counts, rates, and 95\% Wilson confidence intervals.

\subsection{Interpretability framework}

We used SHapley Additive exPlanations (SHAP) to examine how each feature contributes to the Random Forest predictions. Random Forest was selected for the interpretability analysis based on its performance during training-data validation rather than its performance on the held-out test set. Global SHAP importance summarizes the mean absolute contribution of each feature across the held-out windows. The beeswarm plot shows the direction and magnitude of each window-level contribution together with the corresponding feature value.

Because the final document score is the median of its window probabilities, for the local examples we explain the window with a probability closest to the document median. These examples therefore show how the features contribute to the prediction for a representative window of the document rather than providing an explanation of the entire multi-window
aggregation.

\section{Results and discussion}
\label{sec:results}
\subsection{Detection performance}
\begin{figure}[htp]
\centering
\includegraphics[width=2.85in, height=2in]{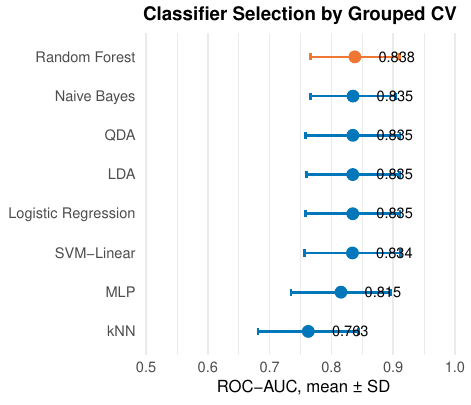}
\caption{Comparison of the eight classifiers using repeated 10-fold cross-validation on the training data, with all data from the same participant kept in the same fold. Points show mean validation ROC-AUC, and error bars show one standard deviation.}
\label{fig:classifier-comparison}
\end{figure}

Figure~\ref{fig:classifier-comparison} shows the cross-validation results used to compare the eight classifiers on the training data. Random Forest achieved the highest mean validation ROC-AUC ($0.838$), followed closely by Naive Bayes, QDA, LDA, Logistic Regression, and linear SVM. The error bars show substantial overlap, indicating similar validation performance
among these classifiers. Random Forest also ranked first in the additional nested-validation check and was therefore selected for the subsequent interpretability analysis.

The classifier comparison and hyperparameter tuning used only the training data, and the held-out test set played no role in these decisions. Final evaluation was performed on 36 documents from 18 held-out participants, using the median of the window-level probabilities to obtain a score for each document. Random Forest achieved a holdout ROC-AUC of $0.870$
(95\% CI: $0.747$--$0.981$) and an F1-score of $0.842$ (95\% CI: $0.737$--$0.944$). Confidence intervals were estimated using
participant-level bootstrap resampling.

\begin{table}[htp]
\centering
\caption{Random Forest performance on 36 held-out documents from 18 unseen participants. GPT-assisted writing is treated as the positive class.}
\label{tab:holdout-performance}
\renewcommand{\arraystretch}{1.15}
\small
\begin{tabular}{lccc}
\toprule
\textbf{Measure} & \textbf{Estimate} & \textbf{Count} & \textbf{95\% CI} \\
\midrule
ROC-AUC        & 0.870  & --   & 0.747--0.981 \\
F1             & 0.842  & --   & 0.737--0.944 \\
False Positive & 22.2\% & 4/18 & 9.0--45.2\% \\
False Negative & 11.1\% & 2/18 & 3.1--32.8\% \\
\bottomrule
\end{tabular}
\end{table}

\begin{table}[htp]
\centering
\caption{Document-level confusion matrix for Random Forest on the 36 held-out documents using the median window-probability rule.}
\label{tab:confusion-matrices}
\renewcommand{\arraystretch}{1.2}
\small
\setlength{\tabcolsep}{3pt}
\begin{tabular}{lcc}
\toprule
 & \textbf{Predicted Human} & \textbf{Predicted GPT} \\
\midrule
\textbf{Actual Human} & \textbf{14} & 4 \\
\textbf{Actual GPT}    & 2 & \textbf{16} \\
\bottomrule
\end{tabular}
\end{table}

With GPT-assisted writing treated as the positive class, four independently authored documents were incorrectly classified as GPT-assisted, resulting in a False Positive Rate of $4/18=22.2\%$ (95\% Wilson CI: $9.0\%$--$45.2\%$). Two GPT-assisted documents were incorrectly classified as independently authored, resulting in a False Negative Rate of $2/18=11.1\%$ (95\% Wilson CI: $3.1\%$--$32.8\%$). The false positives are particularly important in an academic integrity setting because they represent independently authored work incorrectly flagged as GPT-assisted. Both confidence intervals are wide, reflecting the small held-out sample, and these results do not support using the model as standalone evidence for academic misconduct.

As a post-hoc sensitivity analysis, we conducted
additional experiments using five train-test ratios
ranging from 60/40 to 80/20. As shown in Table~\ref{tab:split-sensitivity},
the mean False Positive Rate ranged from
$12.4\%$ to $15.8\%$, while the mean False Negative
Rate ranged from $23.7\%$ to $24.8\%$. Although the mean error rates remained relatively
stable across the evaluated train-test ratios,
the class-specific estimates differed from those
obtained on the original held-out test set.
This variation highlights the uncertainty associated
with evaluation on a single small test set.

\begin{table}[htp]
\centering
\caption{Sensitivity of error rates to train-test
partitioning. Values are means with 2.5th--97.5th
percentile ranges across splits.}
\label{tab:split-sensitivity}
\small
\setlength{\tabcolsep}{3pt}
\begin{tabular}{lcc}
\toprule
\textbf{Train/Test} &
\textbf{FPR} &
\textbf{FNR} \\
\midrule
60/40 & .158 [.083--.244] & .243 [.139--.389] \\
65/35 & .152 [.062--.274] & .242 [.125--.399] \\
70/30 & .142 [.037--.288] & .237 [.082--.370] \\
75/25 & .124 [.045--.262] & .248 [.091--.409] \\
80/20 & .143 [.000--.278] & .244 [.056--.487] \\
\bottomrule
\end{tabular}
\end{table}

\subsection{Error analysis}

The median-probability rule misclassified six of the 36 held-out documents. Four independently authored documents received median GPT-assisted probabilities ranging from $0.636$ to $0.920$ and were therefore false positives. Two GPT-assisted documents were false negatives. One was close to the decision threshold, with a median GPT-assisted probability of $0.495$, while the other received a substantially lower probability of $0.151$. These errors show that the median aggregation reduces the influence of unusually high or low window probabilities but does not eliminate overlap between independently authored and GPT-assisted writing. The results further support using the model as a decision-support tool requiring contextual review rather than as standalone evidence of academic misconduct.

\subsection{Length and feature controls}

Our sliding-window design standardizes the amount of text used to compute the stylometric features while also producing multiple samples from each document. Each document was divided into 250-word windows with a 125-word stride, and the nine features were computed separately for each window. Thus, for most samples, features such as Type-Token Ratio and Hapax Ratio are computed from the same number of words rather than from complete documents of different lengths. Of the 722 windows, 508 contained 250 words. The remaining 214 came from documents shorter than 250 words and
were retained without padding.

We nevertheless examined document length because the original documents differed between the two writing conditions. Independently authored documents contained more words than GPT-assisted documents (paired $t(89)=6.46$, $p<.001$; Welch $p<.001$), with independently authored documents being longer in 72 of the 90 pairs. In contrast, sentence count and mean sentence length did not differ significantly between the two conditions (paired $p=.051$ and $p=.97$, respectively).

As an additional control, a classifier using only the total document word count achieved a cross-validation ROC-AUC of $0.68$, showing that document length itself contains some information about the writing condition. However, total document word count is not used as a feature in our main classifier, and the fixed-word windowing design largely prevents the stylometric features from directly capturing differences in overall document length. Replacing Type-Token Ratio and Hapax Ratio with moving-average TTR produced an ROC-AUC close to that of the full model. Feature-group ablations further showed that lexical features carried most of the discriminative signal, followed by part-of-speech features, while Sentence-Length Variability alone performed near chance.

\subsection{Feature differences}

Table~\ref{tab:feature-tests} shows significant differences in several stylometric features between independently authored and GPT-assisted writing. GPT-assisted documents have higher TTR, Hapax Ratio, Word Entropy, NSR, Noun Ratio, Verb Ratio, and Adjective Ratio, while independently authored documents have higher Adverb Ratio and Sentence-Length Variability. NSR and Noun Ratio show the strongest positive associations with GPT-assisted writing, while Adverb Ratio shows the strongest negative association. For these statistical comparisons, each feature was summarized by its median across the windows of each document, preventing overlapping windows from inflating the inferential sample size.

\subsection{SHAP analysis}
\begin{figure}[htp]
\centering
\includegraphics[width=2.4in, height=2.3in]{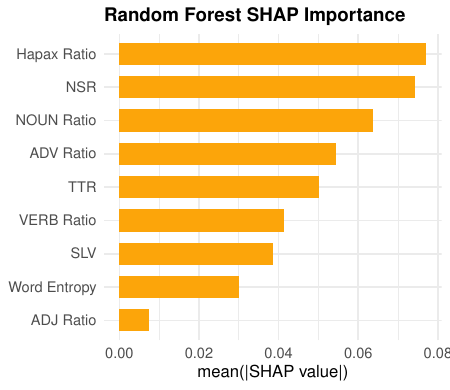}
\caption{Global Random Forest SHAP importance across held-out windows. Higher mean absolute SHAP values indicate greater influence on the predicted probability of GPT-assisted writing.}
\label{fig:shap-importance}
\end{figure}

\begin{figure}[htp]
\centering
\includegraphics[width=2.9in, height=2.8in]{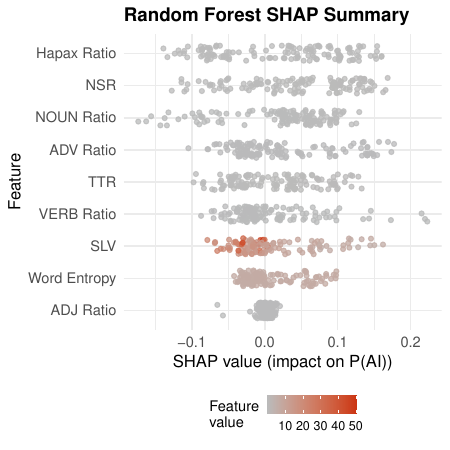}
\caption{Random Forest SHAP summary across held-out windows. Horizontal position shows the contribution to the predicted probability of GPT-assisted writing, and color indicates the corresponding feature value.}
\label{fig:shap-beeswarm}
\end{figure}

Figure~\ref{fig:shap-importance} shows the global Random Forest SHAP
importance. Hapax Ratio is the most influential feature, followed by NSR, Noun Ratio, Adverb Ratio, and TTR. Verb Ratio, Sentence-Length Variability, and Word Entropy have smaller contributions, while Adjective Ratio contributes least. Figure~\ref{fig:shap-beeswarm} shows that the direction and magnitude of these contributions vary across windows, reflecting the non-linear relationships learned by Random Forest.

\begin{figure}[htp]
\centering
\includegraphics[width=3.0in, height=2.6in]{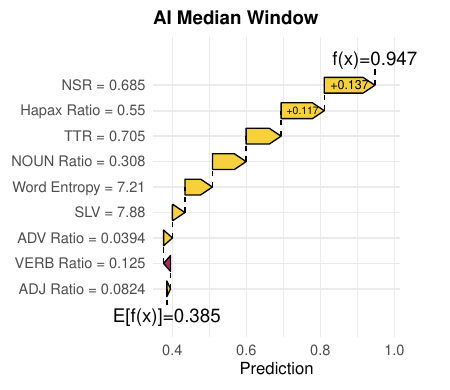}
\caption{SHAP explanation for a representative window of a correctly
classified GPT-assisted document. The selected window has a predicted
probability closest to the document median.}
\label{fig:shap-waterfall-ai}
\end{figure}

\begin{figure}[htp]
\centering
\includegraphics[width=3.0in, height=2.6in]{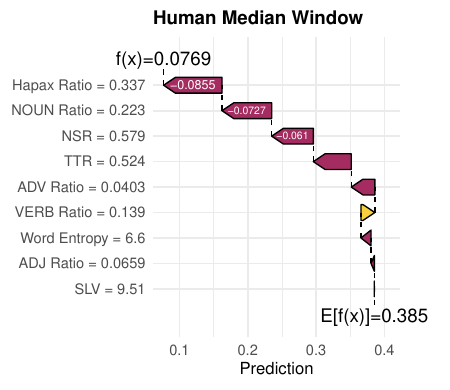}
\caption{SHAP explanation for a representative window of a correctly classified independently authored document. The selected window has a predicted probability closest to the document median.}
\label{fig:shap-waterfall}
\end{figure}

For the GPT-assisted example, the selected window has a predicted
probability of $0.947$, with NSR and Hapax Ratio making the largest positive contributions. For the independently authored example, the selected window has a predicted probability of $0.077$, with Hapax Ratio, Noun Ratio, NSR, and TTR shifting the prediction toward independently authored writing. These examples show how observable stylometric features contribute to individual window predictions. They do not explain the entire multi-window aggregation or establish that a prediction is valid evidence of misconduct.

\section{Ethical considerations and limitations}
\label{sec:ethical}

GPT-assisted writing detection may influence grades and disciplinary actions. Detection systems should therefore be used as decision-support tools rather than definitive evidence of misconduct. A prediction should initiate contextual review and should not replace an instructor's assessment, student testimony, or institutional due process.

False positives are particularly important because they represent
independently authored work incorrectly classified as GPT-assisted. In our held-out evaluation, the False Positive Rate was $22.2\%$ ($4/18$), with a 95\% confidence interval of $9.0\%$--$45.2\%$. The wide interval reflects the small test set and does not support automated use. SHAP can explain which features contributed to a prediction, but it does not establish that the prediction is reliable or fair.

Several limitations bound our findings. The study uses 90 participants from one course, institution, topic area, and GPT-assistance setting. In this dataset, participants used GPT-generated content and were instructed to paraphrase it before typing their responses. Other forms of AI assistance, such as brainstorming or editing, were not evaluated. The results also do not establish performance across other assignments, disciplines, institutions, languages, student populations, or LLMs. Because all participants completed the independent session first, the writing condition may also be affected by session order. External validation is therefore necessary.

Finally, we did not evaluate performance across demographic or language background groups and therefore cannot draw conclusions about fairness across these groups. Writing styles and GPT outputs may also change over time, and the stylometric features may be deliberately altered. Evaluation on new populations and writing conditions is needed before institutional use.

\section{Conclusion and future work}
\label{sec:conclusion}

This study evaluated an interpretable stylometric approach for
distinguishing GPT-assisted and independently authored student writing. Documents were divided into overlapping 250-word windows, and all data from the same participant were kept in the same training, validation, or test set. Eight classifiers were compared using the training data, with Random Forest achieving the highest validation ROC-AUC. Window-level probabilities were then combined using their median to obtain the final document-level prediction.

On 36 held-out documents, Random Forest achieved an ROC-AUC of $0.870$ and an F1-score of $0.842$. Four independently authored documents were incorrectly classified as GPT-assisted, while two GPT-assisted documents were incorrectly classified as independently authored. SHAP identified Hapax Ratio, NSR, Noun Ratio, Adverb Ratio, and TTR as the most influential features. These results show that interpretable stylometric features provide useful signal for distinguishing the two writing conditions. However, the small test set, observed errors, and wide confidence intervals do not support using the model as standalone evidence of academic misconduct.

Future work should evaluate the approach on other institutions, writing tasks, student populations, and forms of AI assistance. It should also examine performance across different groups and determine how much text is needed for reliable classification. Future studies can also evaluate other ways of combining window predictions and combine stylometric features with behavioral information when writing-process data are available.

\printbibliography
\end{document}